%% file: main.tex
\documentclass{article}

\usepackage{microtype}
\usepackage{graphicx}
\usepackage{subfigure}
\usepackage{algorithm}
\usepackage{algorithmic}
\usepackage{booktabs} 

\usepackage{hyperref}

\usepackage[accepted]{icml2020}

\icmltitlerunning{Weight Initializations, Data Orders, and Early Stopping}

\usepackage{mwe,tikz}

\newif\ifcomments
\ifcomments
    \newcommand\gamaga[1]{[\textcolor{blue}{GI: {#1}}]}
    \newcommand\roy[1]{[\textcolor{red}{RS: {#1}}]}
    \newcommand\jdcomment[1]{[\textcolor{teal}{JD: {#1}}]}
    \newcommand\nascomment[1]{[\textcolor{olive}{Noah: {#1}}]}
    \newcommand\hanna[1]{[\textcolor{purple}{Hanna: {#1}}]}

\else
    \providecommand{\gamaga}[1]{}
    \providecommand{\roy}[1]{}
    \providecommand{\jdcomment}[1]{}
    \providecommand{\nascomment}[1]{}
    \providecommand{\hanna}[1]{}
\fi

\begin{document}

\twocolumn[
\icmltitle{Fine-Tuning Pretrained Language Models:\\Weight Initializations, Data Orders, and Early Stopping}



\icmlsetsymbol{equal}{*}

\begin{icmlauthorlist}
\icmlauthor{Jesse Dodge}{cmu,ai2}
\icmlauthor{Gabriel Ilharco}{uw}
\icmlauthor{Roy Schwartz}{ai2,uw}
\icmlauthor{Ali Farhadi}{ai2,uw,xnor}
\icmlauthor{Hannaneh Hajishirzi}{ai2,uw}
\icmlauthor{Noah Smith}{ai2,uw}
\end{icmlauthorlist}

\icmlaffiliation{cmu}{Language Technologies Institute, School of Computer Science, Carnegie Mellon University}
\icmlaffiliation{ai2}{Allen Institute for Artificial Intelligence}
\icmlaffiliation{uw}{Paul G. Allen School of Computer Science and Engineering, University of Washington}
\icmlaffiliation{xnor}{XNOR.AI}

\icmlcorrespondingauthor{Jesse Dodge}{jessed@cs.cmu.edu}

\icmlkeywords{Machine Learning, ICML}

\vskip 0.3in
]



\printAffiliationsAndNotice{}  

\begin{abstract}
Fine-tuning pretrained contextual word embedding models to supervised downstream tasks has become commonplace in natural language processing.
This process, however, is often brittle: even with the same hyperparameter values, distinct random seeds can lead to substantially different results.
To better understand this phenomenon, we experiment with four datasets from the GLUE benchmark, fine-tuning BERT hundreds of times on each while varying only the random seeds.
We find substantial performance increases compared to previously reported results, and we quantify how the performance of the best-found model varies as a function of the number of fine-tuning trials.
Further, we examine two factors influenced by the choice of random seed: weight initialization and training data order. We find that both contribute comparably to the variance of out-of-sample performance, and that some weight initializations perform well across all tasks explored.
On small datasets, we observe that many fine-tuning trials diverge part of the way through training, and we offer best practices for practitioners to stop training less promising runs early.
We publicly release all of our experimental data, including training and validation scores for 2,100 trials, to encourage further analysis of training dynamics during fine-tuning.
\end{abstract}

\input{text/00_intro.tex}
\input{text/01_experimental_setup.tex}
\input{text/02_expected_valid.tex}
\input{text/03_wi_vs_do.tex}
\input{text/04_wi_vs_do_best_vs_worst.tex}
\input{text/05_early_stopping.tex}

\input{text/97_related_work.tex}
\input{text/98_conclusion.tex}

\bibliography{example_paper}
\bibliographystyle{icml2020}

\input{text/99_appendix.tex}

\end{document}

%% file: text/00_intro.tex
\section{Introduction}
\label{sec:intro}

\begin{table}
\setlength\tabcolsep{1.5pt}
    \centering
    \begin{tabular}{l@{\hskip .2in}cccc}
             & MRPC & RTE & CoLA & SST \\\hline
    BERT \cite{phang2018sentence} & 90.7 & 70.0 & 62.1 & 92.5\\
    BERT \cite{liu2019roberta} & 88.0 & 70.4 & 60.6 & 93.2\\
    BERT (ours) & \underline{\textbf{91.4}} & \textbf{77.3} & \textbf{67.6} & \textbf{95.1}\\
    \hline
    STILTs \cite{phang2018sentence} & 90.9 & 83.4 & 62.1 & 93.2 \\
    XLNet \cite{yang2019xlnet} & 89.2 & 83.8 & 63.6 & 95.6\\
    RoBERTa  \cite{liu2019roberta} & 90.9 & 86.6 & 68.0 & 96.4\\
    ALBERT \cite{lan2019albert} & 90.9 & \underline{89.2} & \underline{71.4} & \underline{96.9}\\
    \hline
    \end{tabular}
    \caption{Fine-tuning BERT multiple times while varying only random seeds leads to substantial improvements over previously published validation results with the same model and experimental setup (top rows), on four tasks from the GLUE benchmark. On some tasks, BERT even becomes competitive with more modern models (bottom rows). Best results with standard BERT fine-tuning regime are indicated in bold, best overall results are underscored.
    }
    
    \label{tab:best_perf}
\end{table}

The advent of large-scale self-supervised pretraining has contributed greatly to progress in natural language processing  \cite{devlin2018bert,liu2019roberta,radford2019language}. 
In particular, BERT \cite{devlin2018bert} advanced accuracy on natural language understanding tasks in popular NLP benchmarks such as GLUE \cite{wang2018glue} and SuperGLUE \cite{wang2019superglue}, and variants of this model have since seen adoption in ever-wider applications \cite{schwartz2019bias,lu2019vilbert}.
Typically, these models are first pretrained on large corpora, then fine-tuned on downstream tasks by reusing the model's parameters as a starting point, while adding one task-specific layer trained from scratch. 
Despite its simplicity and ubiquity in modern NLP, this process has been shown to be brittle \cite{devlin2018bert, phang2018sentence, zhu2019freelb, raffel2019exploring}, where fine-tuning performance can vary substantially across different training episodes, even with fixed hyperparameter values.

In this work, we investigate this variation by conducting a series of fine-tuning experiments on four tasks in the GLUE benchmark \cite{wang2018glue}. Changing only training data order and the weight initialization of the fine-tuning layer---which contains only 0.0006\% of the total number of parameters in the model---we find substantial variance in performance across trials. 

We explore how validation performance of the best found model varies with the number of fine-tuning experiments, finding that, even after hundreds of trials, performance has not fully converged.
With the best found performance across all the conducted experiments of fine-tuning BERT, we observe substantial improvements compared to previous published work with the same model (Table \ref{tab:best_perf}).
On MRPC \citep{dolan2005automatically}, BERT performs better than more recent models such as XLNet \cite{yang2019xlnet}, RoBERTa \cite{liu2019roberta} and ALBERT \cite{lan2019albert}. Moreover, on RTE \cite{wang2018glue} and CoLA \cite{warstadt2018neural}, we observe a 7\% (absolute) improvement over previous results with the same model. It is worth highlighting that in our experiments only random seeds are changed---never the fine-tuning regime, hyperparameter values, or pretrained weights.
These results demonstrate how model comparisons that only take into account reported performance in a benchmark can be misleading, and  serve as a reminder of the value of more rigorous reporting practices \cite{dodge-etal-2019-show}.

To better understand the high variance across fine-tuning episodes, we separate two factors that affect it: the weight initialization for the task-specific layer; and the training data order resulting from random shuffling.
The contributions of each of these have previously been conflated or overlooked, even by works that recognize the importance of multiple trials or random initialization \cite{phang2018sentence}.
By conducting experiments with multiple combinations of random seeds that control each of these factors, we quantify their contribution to the variance across runs. 
Moreover, we present evidence that some seeds are consistently better than others in a given dataset for both weight initializations and data orders.
Surprisingly, we find that some weight initializations perform well across \textit{all studied tasks}.

By frequently evaluating the models through training, we empirically observe that worse performing models can often be distinguished from better ones early in training, motivating investigations of early stopping strategies. We show that a simple early stopping algorithm (described in Section \ref{sec:early-stopping}) is an effective strategy for reducing the computational resources needed to reach a given validation performance and include practical recommendations for a wide range of computational budgets.

To encourage further research in analyzing training dynamics during fine-tuning, we publicly release all of our experimental data. This includes, for each of the 2,100 fine-tuning episodes, the training loss at every weight update, and validation performance on at least 30 points in training.

Our main contributions are:
\begin{itemize}
    \item We show that running multiple trials with different random seeds can lead to substantial gains in performance on four datasets from the GLUE benchmark. Further, we present how the performance of the best-found model changes as a function of the number of trials.
    \item We investigate weight initialization and training data order as two sources of randomness in fine-tuning by varying random seeds that control them, finding that 1) they are comparable as sources of variance in performance; 2) in a given dataset, some data orders and weight initializations are consistently better than others; and 3) some weight initializations perform well across multiple different tasks.
    \item We demonstrate how a simple early stopping algorithm can effectively be used to improve expected performance using a given computational budget.
    \item We release all of our collected data of 2,100 fine-tuning episodes on four popular datasets from the GLUE benchmark to incentivize further analyses of fine-tuning dynamics.
\end{itemize}

%% file: text/01_experimental_setup.tex
\section{Methodology}

Our experiments consist of fine-tuning pretrained BERT to four downstream tasks from the GLUE benchmark.
For a given task, we experiment multiple times with the same model using the same hyperparameter values, while modifying only the random seeds that control weight initialization (WI) of the final classification layer and training data order (DO).
In this section we describe in detail the datasets and settings for our experiments.

\subsection{Data}

We examine four datasets from the GLUE benchmark, described below and summarized in Table \ref{tab:datasets}. The data is publicly available and can be download from the repository jiant.\footnote{\url{https://github.com/nyu-mll/jiant}}
Three of our datasets are relatively small (MRPC, RTE, and CoLA), and one relatively large (SST).
Since all datasets are framed as binary classification, the model structure for each is the same, as only a single classification layer with two output units is appended to the pretrained BERT.

\label{sec:data}
\paragraph{Microsoft Research Paraphrase Corpus} (MRPC; \citealp{dolan2005automatically}) contains pairs of sentences, labeled as either nearly semantically equivalent, or not. The dataset is evaluated using the average of $F_1$ and accuracy.
\paragraph{Recognizing Textual Entailment} (RTE;   \citealp{wang2018glue}) combines data from a series of datasets \cite{dagan2005pascal, bar2006second, giampiccolo2007third, bentivogli2009fifth}. Each example in RTE is a pair of sentences, and the task is to predict whether the first (the premise) entails the second (the hypothesis).
\paragraph{Corpus of Linguistic Acceptability} (CoLA;
 \citealp{warstadt2018neural}) is comprised of English sentences labeled as either grammatical or ungrammatical. Models are evaluated on Matthews correlation (MCC; \citealp{matthews1975comparison}), which ranges between --1 and 1, with random guessing being 0.
\paragraph{Stanford Sentiment Treebank} (SST; \citealp{socher2013recursive}) consists of sentences annotated as expressing \textit{positive} or \textit{negative} sentiment (we use the binary version of the annotation), collected from movie reviews.

\begin{table}
\setlength\tabcolsep{2.8pt}
    \centering
    \begin{tabular}{lcccc}
                                  & MRPC & RTE & CoLA & SST\\\hline
    evaluation metric             & Acc./$F_1$ & Acc. & MCC & Acc.\\ 
    majority baseline       & 0.75 & 0.53 & 0.00 & 0.51 \\
    \hline
    \# training samples           & 3.7k & 2.5k & 8.6k & 67k\\
    \# validation samples         & 409 & 277 & 1,043 & 873\\
    \end{tabular}
    \caption{The datasets used in this work, which comprise four out of nine of the tasks in the GLUE benchmark \cite{wang2018glue}.}
    \label{tab:datasets}
\end{table}

\subsection{Fine-tuning}
\label{sec:ft}

Following standard practice, we fine-tune BERT (BERT-large, uncased) for three epochs \cite{phang2018sentence, devlin2018bert}. We fine-tune the entire model (340 million parameters), of which the vast majority start as pretrained weights and the final layer (2048 parameters) is randomly initialized. The weights in the final classification layer are initialized using the standard approach used when fine-tuning pretrained transformers like BERT, RoBERTa, and ALBERT \cite{devlin2018bert,liu2019roberta,lan2019albert}: sampling from a normal distribution with mean 0 and standard deviation 0.02. All experiments were run on P100 GPUs with 16 GB of RAM. We train with a batch size of 16, a learning rate of 0.00002, and dropout of 0.1; the open source implementation, pretrained weights, and full hyperparameter values and experimental details can be found in the HuggingFace transformer library \cite{Wolf2019HuggingFacesTS}.\footnote{\url{https://github.com/huggingface/transformers}}

Each experiment is repeated $N^2$ times, with all possible combinations of $N$ distinct random seeds for WI and $N$ for DO.\footnote{Although any random numbers would have sufficed, for completeness: we use the numbers \{$1,\dots,N$\} as seeds.}   For the datasets MRPC, RTE, and CoLA, we run a total of $625$ experiments each ($N{=}25$). For the larger SST, we run $225$ experiments ($N{=}15$).

%% file: text/02_expected_valid.tex
\section{The large impact of random seeds}
\label{sec:expected_valid}




\begin{figure*}
  \centering   
  \includegraphics[width=\linewidth]{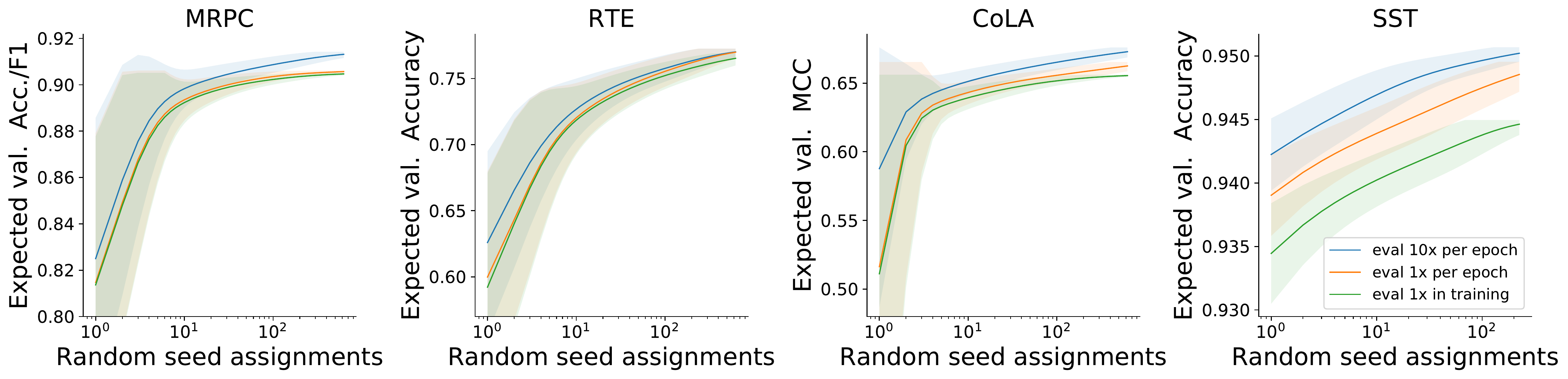}
  \caption{Expected validation performance \cite{dodge-etal-2019-show}, plus and minus one standard deviation, as the number of experiments increases. The $x$-axis represents the budget (e.g., $x=10$ indicates a budget large enough to train 10 models). The $y$-axis is the expected performance of the best of the $x$ models trained. Each plot shows three evaluation scenarios: in the first, the model is frequently evaluated on the validation set during training (blue); in the second, at the end of each epoch (orange); and in the third, only at the end training (green). As we increase the number of evaluations per run we see higher expected performance and smaller variances. Further, more frequently evaluating the model on validation data leads to higher expected validation values.}
\label{fig:expected_valid}
\end{figure*}

Our large set of fine-tuning experiments evidences the sizable variance in performance across trials varying only random seeds.
This effect is especially pronounced on the smaller datasets;
the validation performance of the best-found model from multiple experiments is substantially higher than the expected performance of a single trial.
In particular, in Table~\ref{tab:best_perf} we report the performance of the best model from all conducted experiments, which represents substantial gains compared to previous work that uses the same model and optimization procedure.
On some datasets, we observe numbers competitive with more recent models which have improved pretraining regimes \cite{phang2018sentence, yang2019xlnet, liu2019roberta, lan2019albert}; compared to BERT, these approaches pretrain on more data, and some utilize more sophisticated modeling or optimization strategies.
We leave it to future work to analyze the variance from random seeds on these other models, and note that running analogous experiments would likely also lead to performance improvements.

In light of these overall gains and the computational burden of running a large number of experiments, we explore how the number of trials influences the expected validation performance.

\subsection{Expected validation performance} 

To quantify the improvement found from running more experiments, we turn to expected validation performance as introduced by \citet{dodge-etal-2019-show}. The standard machine learning experimental setup involves a practitioner training $x$ models, evaluating each of them on validation data, then taking the model which has the best validation performance and evaluating it on test data.
Intuitively, as the number of trained models $x$ increases, the best of those $x$ models will improve; expected validation performance calculates the expected value of the best validation performance as a function of $x$.\footnote{A full derivation can be found in \citet{dodge-etal-2019-show}.
}


We plot expected validation curves for each dataset in Figure \ref{fig:expected_valid} with (plus or minus) the standard deviation shaded.\footnote{We shade between the observed minimum and maximum.}
The leftmost point on each of these curves ($x=1$) shows the expected performance for a budget of a single training run.  
For all datasets, Figure \ref{fig:expected_valid} shows, unsurprisingly, that expected validation performance increases as more computational resources are used.
This rising trend continues even up to our largest budget, suggesting even larger budgets could lead to improvements.
On the three smaller datasets (MRPC, RTE, and CoLA) there is significant variance at smaller budgets, which indicates that individual runs can have widely varying performance.

In the most common setup for fine-tuning on these datasets, models are evaluated on the validation data after each epoch, or once after training for multiple epochs \cite{phang2018sentence, devlin2018bert}. 
In Figure \ref{fig:expected_valid} we show expected performance as we vary the number of evaluations on validation data during training (all models trained for three epochs): once after training (green), after each of the three epochs (orange), and frequently throughout training (ten times per epoch, blue).\footnote{Compared to training, evaluation is typically cheap, since the validation set is  smaller than the training set and evaluation requires only a forward pass. Moreover, evaluating on the validation data can be done in parallel to training, and thus does not necessarily slow down training.}  Considering the benefits of more frequent evaluations as shown in Figure \ref{fig:expected_valid}, we thus recommend this practice in similar scenarios.

%% file: text/03_wi_vs_do.tex
\section{Weight initialization and data order}
\label{sec:wi_vs_do}

\begin{figure*}[h!]
  \centering   
  \includegraphics[width=\linewidth]{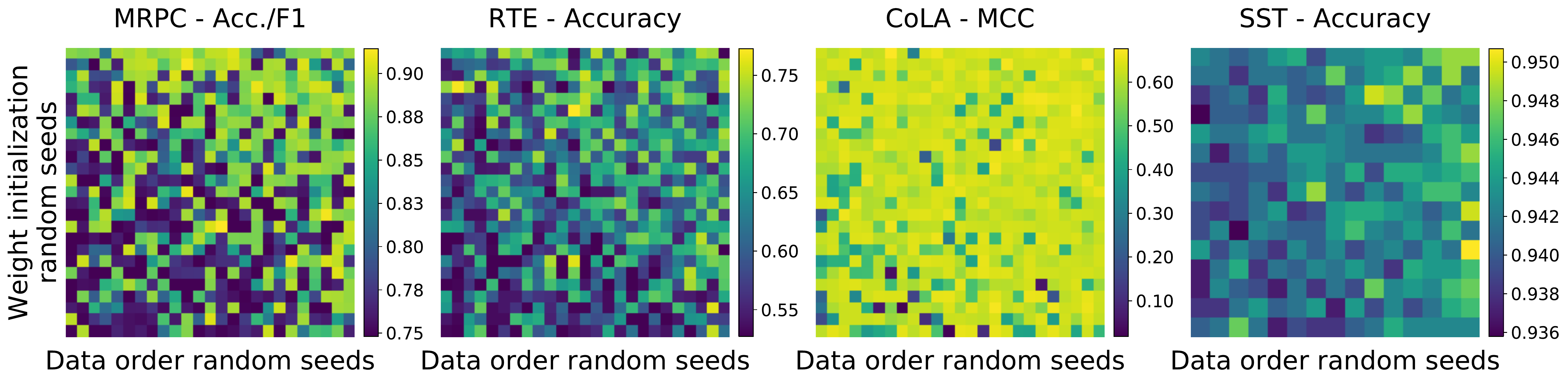}
  \caption{A visualization of validation performance for all experiments, where each colored cell represents the performance of a training run with a specific WI and DO seed. Rows and columns are sorted by their average, such that the best WI seed corresponds to the top row of each plot, and the best DO seed correspond to the right-most column. Especially on smaller datasets a large variance in performance is observed across different seed combinations, and on MRPC and RTE models frequently diverge, performing close to the majority baselines (listed in Table \ref{tab:datasets}).}
\label{fig:matrices}
\end{figure*}

\begin{figure*}[h!]
  \centering   
    \includegraphics[width=\linewidth]{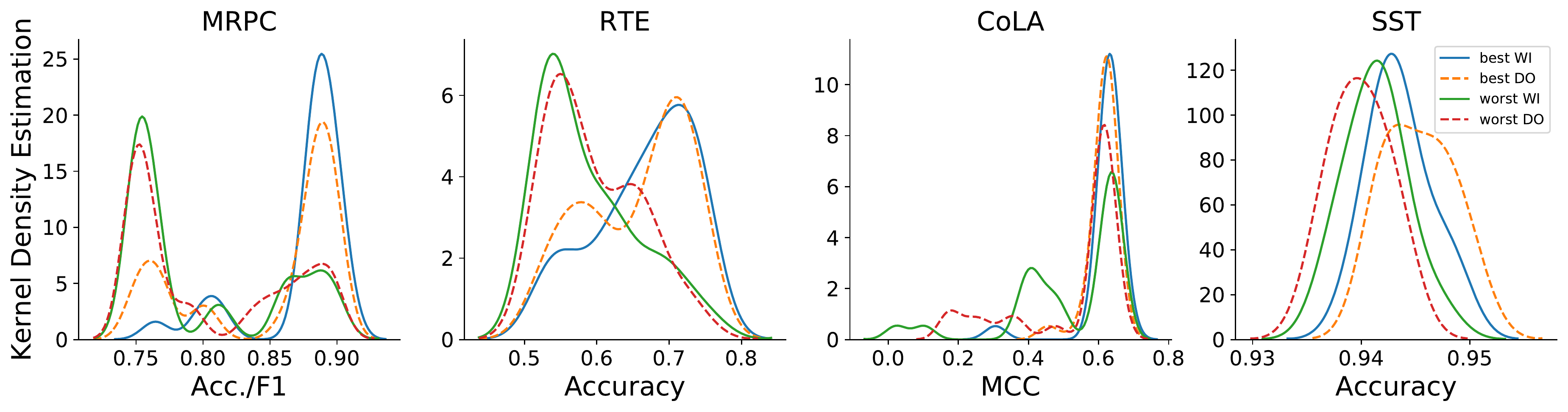}
  \caption{Some seeds are better then others. Plots show the kernel density estimation of the distribution of validation performance for best and worst WI and DO seeds. Curves for DO seeds are shown in dashed lines and for WI in solid lines. MRPC and RTE exhibit pronounced bimodal shapes, where one of the modes represents divergence; models trained with the worst WI and DO are more likely to diverge than learn to predict better than random guessing. Compared to the best seeds, the worst seeds are conspicuously more densely populated in the lower performing regions, for all datasets.}
\label{fig:best_and_worst}
\end{figure*}

To better understand the high variance in performance across trials, we analyze two source of randomness: the weight initialization of the final classification layer and the order the training data is presented to the model. While previous work on fine-tuning pretrained contextual representation models \cite{devlin2018bert, phang2018sentence} has generally used a single random seed to control these two factors, we analyze them separately.

Our experiments are conducted with every combination of a set of weight initialization seeds (WI) and a set of data order (DO) seeds that control these factors. One data order can be viewed as one sample from the set of permutations of the training data. Similarly, one weight initialization can be viewed as a specific set of samples from the normal distribution from which we draw them.


An overview of the collected data is presented in Figure \ref{fig:matrices}, where each colored cell represents the validation performance for a single experiment.
In the plots, each row represents a single weight initialization and each column represents a single data order.
We sort the rows and columns by their averages; the top row contains experiments with the WI with the highest average performance, and the rightmost column contains experiments with the DO with the highest average performance.\footnote{Each cell represents an independent sample, so the rows and columns can be reordered.}

For MRPC, RTE, and CoLA, a fraction of the trained models diverge, yielding performance close to that of predicting the most frequent label (see Table \ref{tab:datasets}).
This partially explains the large variance found in the expected validation curves for those three datasets in Figure \ref{fig:expected_valid}.

\subsection{Decoupling}

 \begin{table}
\setlength\tabcolsep{2.8pt}
    \centering
    \begin{tabular}{lcccc}
                 & MRPC & RTE & CoLA & SST\\\hline
    Agg. over WI & .058 & .066 & .090 & .0028\\
    Agg. over DO & .059 & .067 & .095 & .0024\\\hline
    Total        & .061 & .069 & .101 & .0028\\
    \end{tabular}
    \caption{Expected (average) standard deviation in validation performance across runs. The expected standard deviation of given WI and DO random seeds are close in magnitude, and only slightly below the overall standard deviation.}
    \label{tab:std_agg}
\end{table}

From Figure \ref{fig:matrices}, it is clear that different random seed combinations can lead to substantially different validation performance. In this section, we investigate the sources of this variance, decoupling the distribution of performance based on each of the factors that control randomness.

For each dataset, we compute for each WI and each DO seed the standard deviation in validation performance across all trials with that seed. We then compute the expected (average) standard deviation, aggregated under all WI or all DO seeds, which are shown in Table \ref{tab:std_agg}; we show the distribution of standard deviations in the appendix. Although their magnitudes vary significantly between the datasets, the expected standard deviation from the WI and DO seeds is comparable, and are slightly below the overall standard deviation inside a given task. 

\subsection{Some random seeds are better than others}
To investigate whether some WI or DO seeds are better than their counterparts,  
Figure \ref{fig:best_and_worst} plots the random seeds with the best and worst average performance. 
The best and worst seeds exhibit quite different behavior: compared to the best, the worst seeds have an appreciably higher density on lower performance ranges, indicating that they are generally inferior.
On MRPC, RTE, and CoLA the performance of the best and worst WIs are more dissimilar than the best and worst DOs, while on SST the opposite is true.
This could be related to the size of the data; MRPC, RTE, and CoLA are smaller datasets, whereas SST is larger, so SST has more data to order and more weight updates to move away from the initialization.

Using ANOVA \cite{fisher1935anova} to test for statistical significance, we examine whether the performance of the best and worst DOs and WIs have distributions with different means.
The results are shown in Table \ref{tab:stat_tests}.
For all datasets, we find the best and worst DOs and WIs are significantly different in their expected performance ($p < 0.05$). We include a discussion of the assumptions behind ANOVA in the appendix.

\begin{table}
 \setlength\tabcolsep{3.8pt}
    \centering
        \begin{tabular}{l@{\hskip .2in}cccc}
        & MRPC & RTE & CoLA & SST \\ \hline \\[-2.3ex]
        WI & $2.0{\times}10 {^{-6}}$ & $2.8{\times}10^{-4}$ & $7.0{\times}10^{-3}$ & $3.3{\times}10^{-2}$ \\
        DO & $8.3{\times}10 ^{-3}$ & $3.2{\times}10^{-3}$ & $1.1{\times}10^{-2}$ & $1.3{\times}10^{-5}$ \\
       \end{tabular}
    \caption{
    $p$-values from ANOVA indicate that there is evidence to reject the null hypothesis that the performance of the best and worst WIs and DOs have distributions with the same means ($p < 0.05$).
}
    \label{tab:stat_tests}
\end{table}


%% file: text/04_wi_vs_do_best_vs_worst.tex
\subsection{Globally good initializations} 
A natural question that follows is whether some random seeds are good \emph{across datasets}.
While the data order is dataset specific, the same weight initialization can be applied to multiple classifiers trained with different datasets:
since all tasks studied are binary classification, models for all datasets share the same architecture, including the classification layer that is being randomly initialized and learned. 

\begin{figure*}[h!]
  \centering   
  \includegraphics[width=\linewidth]{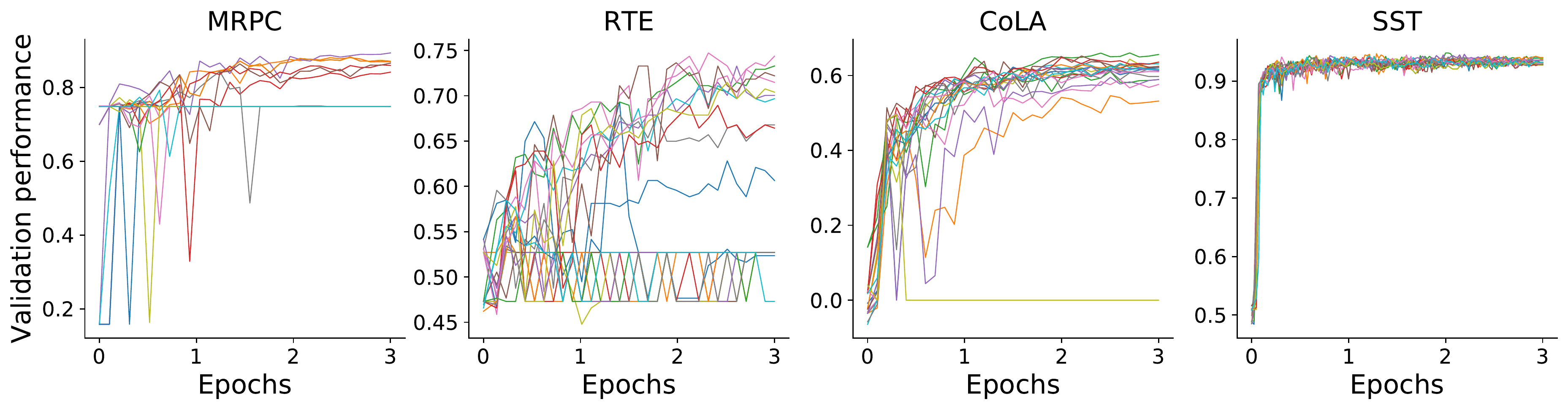}
  \caption{Some promising seeds can be distinguished early in training. The plots show training curves for 20 random WI and DO combinations for each dataset. Models are evaluated every 10th of an epoch (except SST, which was evaluated every 100 steps, equivalent to 42 times per epoch).
  For the smaller datasets, training is unstable, and a non-negligible portion of the models yields poor performance, which can be identified early on.
}
\label{fig:training_curves}
\end{figure*}
\begin{figure*}
  \centering   
  \includegraphics[width=\linewidth]{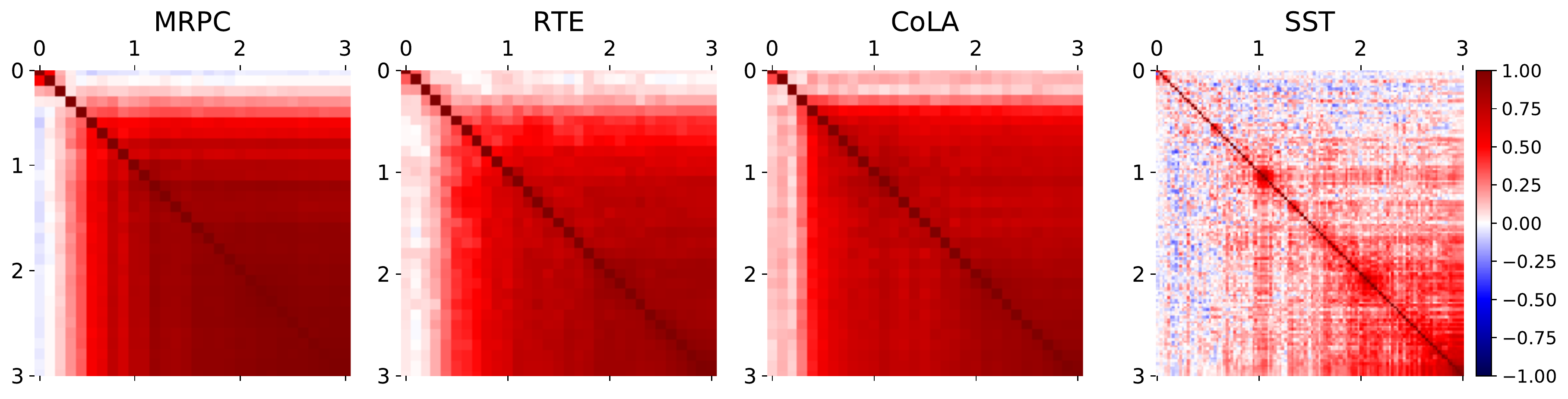}
\caption{Performance early in training is highly correlated with performance late in training. Each figure shows the Spearman's rank correlation between the validation performance at different points in training; the axes represent epochs. A point at coordinates $i$ and $j$ in the plots indicates the correlation between the best found performances after $i$ and after $j$ evaluations. Note that the plots are symmetric.}
\label{fig:correlation}
\end{figure*}

We compare the different weight initializations across datasets.
We find that some initializations perform consistently well. For instance, WI seed 12 has the best performance on CoLA and RTE, the second best on MRPC, and third best on SST. This suggests that, perhaps surprisingly, some weight initializations perform well across tasks.

Studying the properties of good weight initializations and data orders is an important question that could lead to significant empirical gains and enhanced understanding of the fine-tuning process. We defer this question to future work, and release the results of our 2,100 fine-tuning experiments to facilitate further study of this question by the community.



%% file: text/05_early_stopping.tex
%
\section{Early stopping}
\label{sec:early-stopping}


Our analysis so far indicates a high variance in the fine-tuning performance of BERT when using different random seeds, where some models fail to converge.\footnote{This was also observed by \citet{phang2018sentence}, who showed that their proposed STILTs approach reduced the number of diverging  models.} In this section we show that better performance can be achieved with the same computational resources by using early stopping algorithms that stop the least promising trials early in training. We also include recommendations for practitioners for setting up experiments meeting a variety of computational budgets.

\begin{figure}
  \centering   
  \includegraphics[width=.98\linewidth]{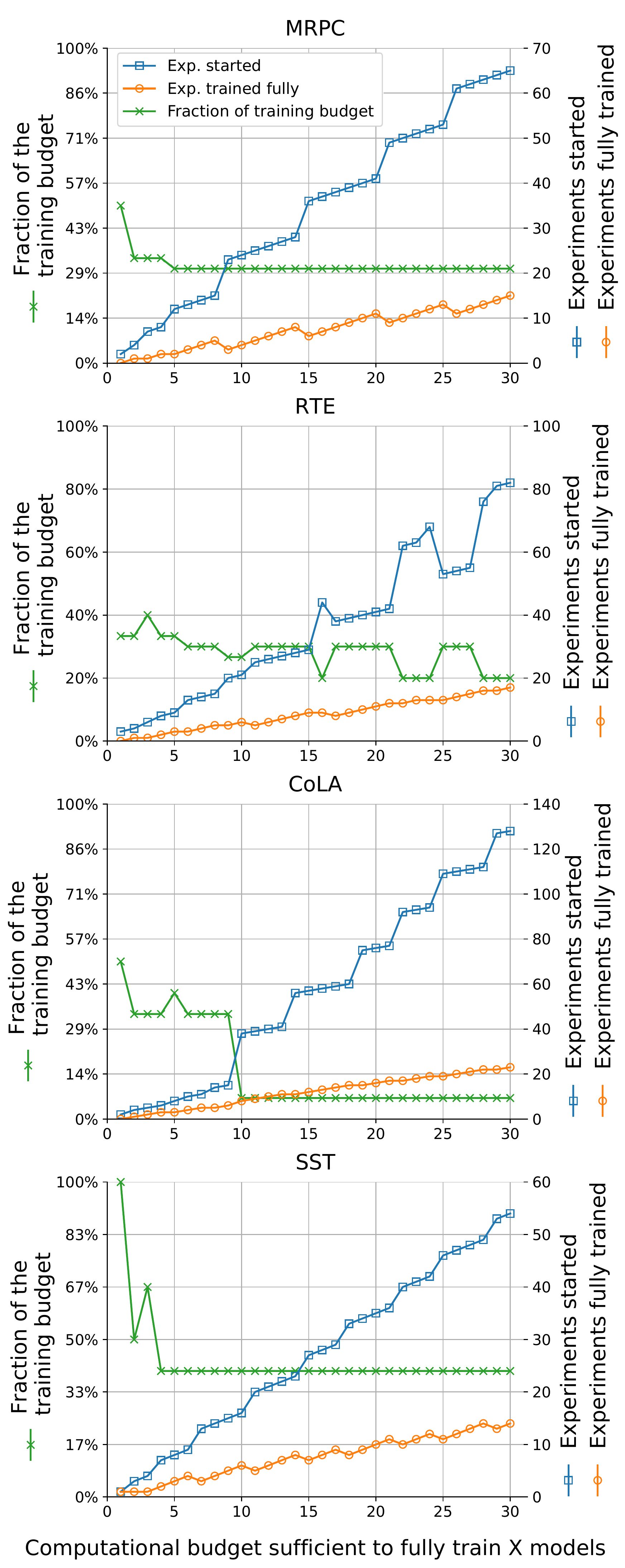}
\caption{Best observed early stopping parameters on each dataset. For a given budget large enough to fully train $x$ models (each trained for 3 epochs), this plot shows the optimal parameters for early stopping. For instance, in MRPC with a budget large enough for 20 trials, the best observed performance came by starting 41 trials (blue), then continuing only the 11 most promising trials (orange) after 30\% of training (green).}
\label{fig:early_stop}
\end{figure}

\paragraph{Early discovery of failed experiments} Figure~\ref{fig:training_curves} shows that performance divergence can often be recognized early in training. These plots show the performance values of 20 randomly chosen models at different times across training. In many of the curves, continuing training of lower performing models all the way through can be a waste of computation. In turn, this suggests the potential of early stopping least promising trials as a viable means of saving computation without large decreases in expected performance. For instance, after training halfway through the first epoch on CoLA the models which diverged could be stopped.


We further examine the correlation of validation performances at different points throughout training, shown in Figure~\ref{fig:correlation}. One point in one of these plots represents the Spearman's rank correlation between performance at iteration $i$ and iteration $j$ across trials. High rank correlation means that the ranking of the models is similar between the two evaluation points, and suggests we can stop the worst performing models early, as they would likely continue to underperform.\footnote{Similar plots with Pearson correlation can be found in the appendix.}
On MRPC, RTE and CoLA, there exists a high correlation between the models' performance early on (part way through the first epoch) and their final performance. On the larger SST dataset, we see high correlation between the performance after training for two epochs and the final performance.

\paragraph{Early stopping} Considering the evidence from the training curves and correlation plots, we analyze a simple algorithm for early stopping.
Our algorithm is inspired by existing approaches to making hyperparameter search more efficient by stopping some of the least promising experiments early \cite{Jamieson:2016,li2016hyperband}.\footnote{``Early stopping'' can also relate to stopping a single training run if the loss hasn't decreased for a given number of epochs. Here we refer to the notion of stopping a subset of  multiple trials.}
Here we apply an early stopping algorithm to select the best performing random seed.\footnote{Our approach does not distinguish between DO and WI. While initial results suggest that this distinction could inspire more sophisticated early-stopping criteria, we defer this to future work.}
The algorithm has three parameters: $t$, $f$, and $p$.
We start by training $t$ trials, and partially through training ($f$, a fraction of the total number of epochs) evaluate all of them and only continue to fully train the $p$ most promising ones, while discarding the rest.
This algorithm takes a total of $(tf+p(1-f))s$ steps, where $s$ is the number of steps to fully train a model.\footnote{In our experiments, $s=3$ epochs.}

\paragraph{Start many, stop early, continue some}  As shown earlier, the computational budget of running this algorithm can be computed directly from an assignment to the parameters $t$, $f$, and $p$. Note that there are different ways to assign these parameters that lead to the same computational budget, and those can lead to significantly distinct performance in expectation; to estimate the performance for each configuration we simulate this algorithm by sampling 50,000 times from from our full set of experiments. In Figure~\ref{fig:early_stop} we show the best observed assignment of these parameters for budgets between 3 and 90 total epochs of training, or the equivalent of 1 to 30 complete training trials. There are some surprisingly consistent trends across datasets and budgets -- the number of trials started should be significantly higher than the number trained fully, and the number of trials to train fully should be around $x/2$. On three out of four datasets, stopping least promising trials after 20--30\% of training (less than one epoch) yielded the best results---and on the fourth dataset this is still a strong strategy.

\begin{figure}
  \centering   
  \includegraphics[width=.9\linewidth]{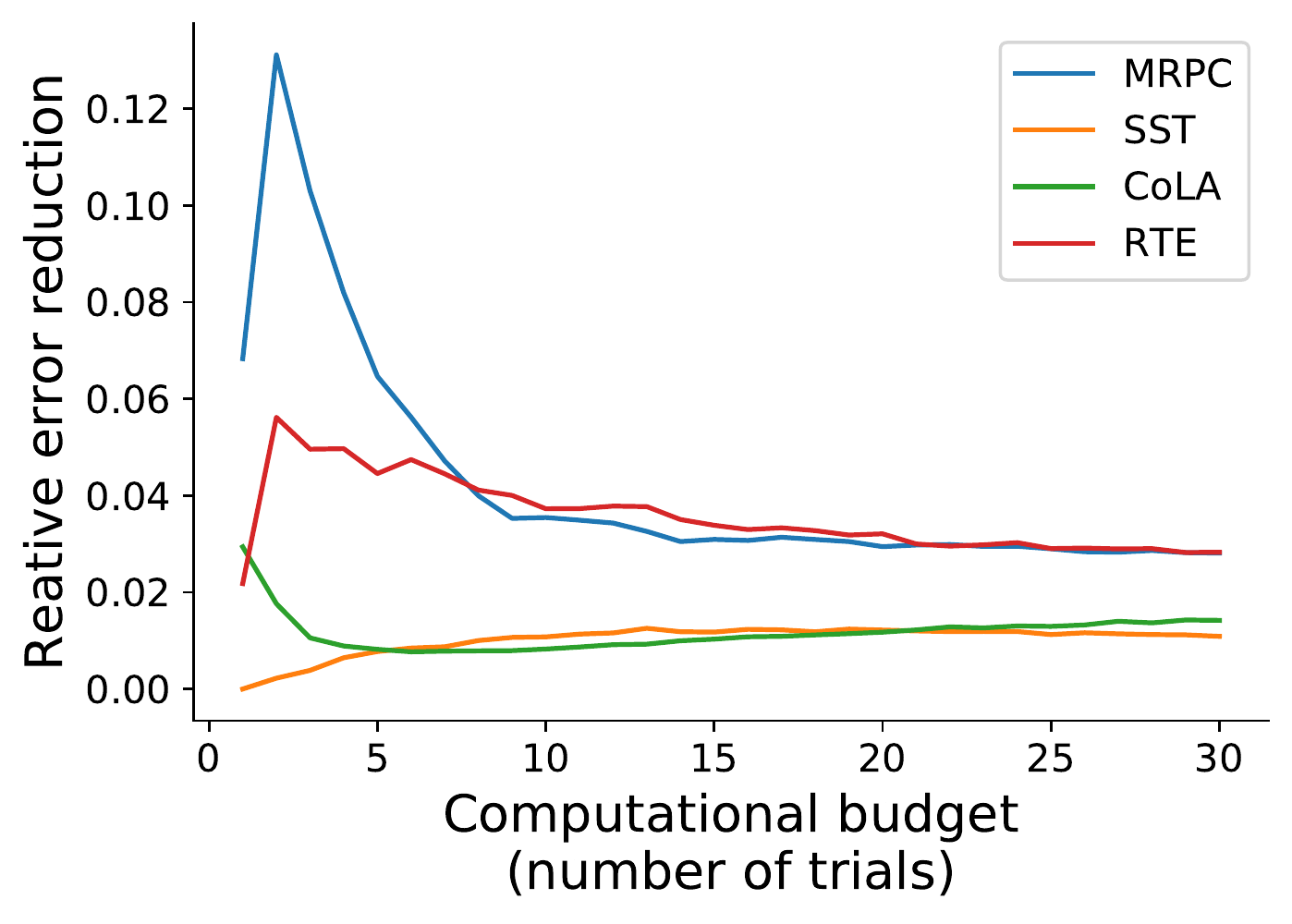}
\caption{Relative error reduction from the early stopping approach in Figure~\ref{fig:early_stop}, compared to the baseline of training $x$ models on the full training budget.
Performance on RTE and SST is measured using accuracy, on MRPC it is the average of accuracy and F1, and on CoLA it is MCC.
``Error'' here refers to one-minus-performance for each of these datasets.
As the budget increases, the absolute performance on all four datasets increases, and the absolute improvement from early stopping is fairly consistent.}
\label{fig:rel_improvement}
\end{figure}

\paragraph{Early stopping works} We compare this algorithm with our baseline of running multiple experiments all the way through training, without any early stopping ($f$=$1$, $t$=$p$) and using the same amount of computation. Specifically, for a given computational budget equivalent to fully training $t$ models, we measure improvement as the relative error reduction from using early stopping with the best found settings for that computational budget.
Figure~\ref{fig:rel_improvement} shows the relative error reduction for each dataset as the computational budget varies, where we observe small but reasonably consistent improvements on all tasks. 

%% file: text/97_related_work.tex
\section{Related work}
\label{sec:related}

Most work on hyperparameter optimization tunes a number of impactful hyperparameters, such as the learning rate, the width of the layers in the model, and the strength of the regularization \cite{li2016hyperband, bergstra2011hparam}.
For modern machine learning models such tuning has proven to have a large impact on the performance; in this work we only examine two oft-overlooked choices that can be cast as hyperparameters and still find room for optimization.

\citet{melis2018lstm} heavily tuned the hyperpamareters of an LSTM language model, for some experiments running 1,500 rounds of Bayesian optimization (thus, training 1,500 models). 
They showed that an LSTM, when given such a large budget for hyperparameter tuning, can outperform more complicated neural models.
While such work informs the community about the best performance found after expending very large budgets, it is difficult for future researchers to build on this without some measure of how the performance changes as a function of computational budget.
Our work similarly presents the best-found performance using a large budget (Table~\ref{tab:best_perf}), but also includes estimates of how performance changes as a function of budget (Figure~\ref{fig:expected_valid}).

A line of research has addressed the distribution from which initializations are drawn. 
The Xavier initialization \cite{glorot2010init} and Kaiming initialization \cite{he2015init} initialize weights by sampling from a uniform distribution or normal distribution with variance scaled so as to preserve gradient magnitudes through backpropagation.
Similarly, orthogonal initializations \cite{saxe2014orthogonal} aim to prevent exploding or vanishing gradients.
In our work, we instead examine how different samples from an initialization distribution behave, and we hope future work which introduces new initialization schemes will provide a similar analysis.

Active learning techniques, which choose a data order using a criterion such as the model's uncertainty \cite{lewis1994active}, have a rich history.
Recently, it has even been shown that that training on mini-batches which are diverse in terms of data or labels \cite{zhang2017diversity} can be more sample efficient.
The tools we present here can be used to evaluate different seeds for a stochastic active learning algorithm, or to compare different active learning algorithms.

%% file: text/98_conclusion.tex
\section{Conclusion}

In this work we study the impact of random seeds on fine-tuning contextual embedding models, the currently dominant paradigm in NLP. We conduct a large set of experiments on four datasets from the GLUE benchmark and observe significant variance across these trials. Overall, these experiments lead to substantial performance gains on all tasks. By observing how the expected performance changes as we allocate more computational resources, we expect that further gains would come from an even larger set of trials. Moreover, we examine the two sources of variance across trials, weight initialization and training data order, finding that in expectation, they contribute comparably to the variance in performance. 
Perhaps surprisingly, we find that some data orders and  initializations are better than others, and the latter can even be observed even across tasks. 
A simple early stopping strategy along with practical recommendations is included to alleviate the computational costs of running multiple trials. All of our experimental data containing thousands of fine-tuning episodes is publicly released.

%% file: text/99_appendix.tex
\clearpage
\newpage
\onecolumn

\appendix

\section{Appendix}
\begin{figure*}[h!]
   \centering   
   \includegraphics[width=\linewidth]{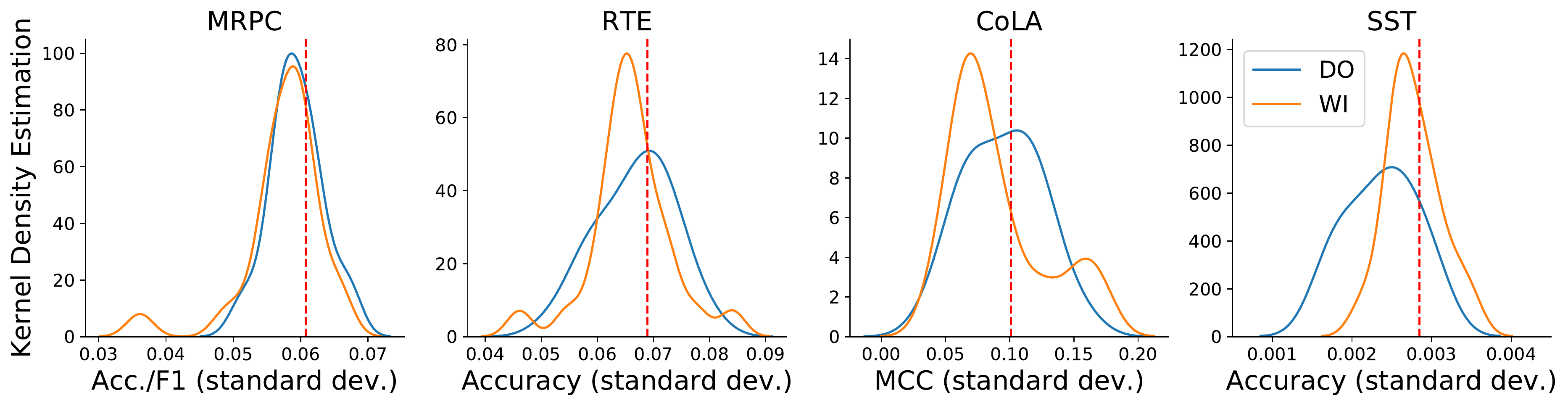}
   \caption{Kernel density estimation of the distribution of standard deviation in validation performance aggregated under fixed random seeds, either for weight initialization (blue) or data order (orange). The red dashed line shows the overall standard deviation for each dataset. The DO and WI curves have expected standard deviation values of similar magnitude, which are also comparable with the overall standard deviation.}
\label{fig:std_kde}
\end{figure*}

We plot the distribution of standard deviations in final validation performance across multiple runs, aggregated under a fixed random seed, either for weight initialization or data order. The results are shown in Figure \ref{fig:std_kde}, indicating that the inter-seed aggregated variances are comparable in magnitude, considering aggregation over both WI and DO.

\section{ANOVA assumptions}
ANOVA makes three assumptions: 1) independence of the samples, 2) homoscedasticity (roughly equal variance across groups), and 3) normally distributed data. 

ANOVA is not robust to violations of independence, but each DO and WI is an I.I.D. sample, and thus independent. 
ANOVA is generally robust to groups with somewhat differing variance if the groups are the same size, which is true in our experiments.
ANOVA is more robust to non-normally distributed data for larger sample sizes; our SST experiments are quite close to normally distributed, and the distribution of performance on the smaller datasets is less like a normal distribution but we have larger sample sizes.

\section{Pearson Correlation}
\begin{figure*}[h!]
  \centering   
  \includegraphics[width=\linewidth]{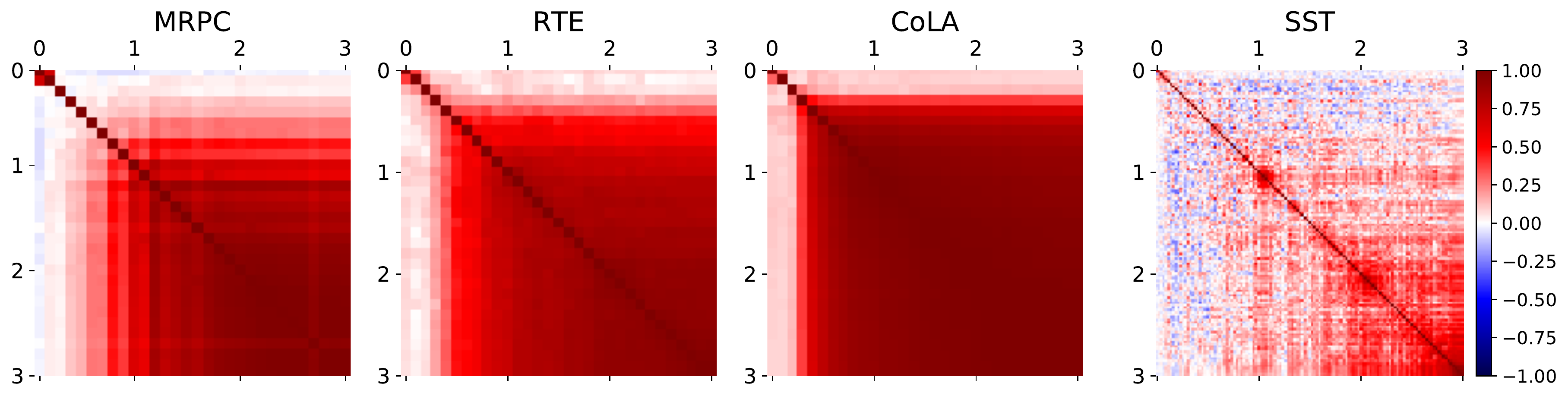}
\caption{Performance early in training is highly correlated with performance late in training. Each figure shows the Spearman's rank correlation between the validation performance at different points in training; the axes represent epochs. A point at coordinates $i$ and $j$ in the plots indicates the correlation between the best found performances after $i$ and after $j$ evaluations. Note that the plots are symmetric.}
\label{fig:pearson_correlation}
\end{figure*}

In Figure~\ref{fig:pearson_correlation} we include the Pearson correlation between different points in training, whereas Figure~\ref{fig:correlation} showed the \emph{rank} correlation of the same data. One point in one of these plots represents the Pearson's correlation between performance at iteration $i$ and iteration $j$ across trials. High correlation means that the performance of the models is similar between the two evaluation points.

